\documentclass[review]{elsarticle}

\usepackage{framed}
\usepackage{graphicx}
\usepackage{booktabs,balance,makecell,bm,caption,hyperref,amsmath,subfigure,stfloats,float,lineno}
\usepackage{amssymb}
\usepackage{bm,color}
\usepackage{multirow,threeparttable,diagbox,lscape,amsmath,algorithm,algpseudocode}
\usepackage[table]{xcolor}
\usepackage{soul}
\newcommand{\rev}{}
\newsavebox\CBox
\def\textBF#1{\sbox\CBox{#1}\resizebox{\wd\CBox}{\ht\CBox}{\textbf{#1}}}

\journal{Expert Systems with Applications}
\biboptions{authoryear}
\begin{document}

\begin{frontmatter}
\begin{titlepage}
\begin{center}
\vspace*{1cm}

\textbf{ \large A Semi-Supervised Adaptive Discriminative Discretization Method Improving Discrimination Power of Regularized Naive Bayes}

\vspace{1.5cm}

Shihe Wang$^{a}$ (Shihe.Wang@nottingham.edu.cn), Jianfeng Ren$^a$ (Jianfeng.Ren@nottingham.edu.cn), Ruibin Bai$^a$ (Ruibin.Bai@nottingham.edu.cn) \\

\hspace{10pt}

\begin{flushleft}
\small  
$^a$ School of Computer Science, University of Nottingham Ningbo China, Ningbo 315100, China\\

\vspace{1cm}
\textbf{Corresponding Author:} \\
Jianfeng Ren \\
School of Computer Science, University of Nottingham Ningbo China, Ningbo 315100, China \\
Email: Jianfeng.Ren@nottingham.edu.cn

\end{flushleft}        
\end{center}
\end{titlepage}

\title{A Semi-Supervised Adaptive Discriminative Discretization Method Improving Discrimination Power of Regularized Naive Bayes}

\author[b]{Shihe Wang}
\ead{Shihe.Wang@nottingham.edu.cn}
\author[b]{Jianfeng Ren\corref{1}}
\ead{Jianfeng.Ren@nottingham.edu.cn}
\cortext[1]{Corresponding author}        
\author[b]{Ruibin Bai}
\ead{Ruibin.Bai@nottingham.edu.cn}
\address[b]{School of Computer Science, University of Nottingham Ningbo China, Ningbo 315100, China}

\begin{abstract}
Recently, many improved naive Bayes methods have been developed with enhanced discrimination capabilities. Among them, regularized naive Bayes (RNB) produces excellent performance by balancing the discrimination power and generalization capability. Data discretization is important in naive Bayes. By grouping similar values into one interval, the data distribution could be better estimated. However, existing methods including RNB often discretize the data into too few intervals, which may result in a significant information loss. To address this problem, we propose a semi-supervised adaptive discriminative discretization framework for naive Bayes, which could better estimate the data distribution by utilizing both labeled data and unlabeled data through pseudo-labeling techniques. The proposed method also significantly reduces the information loss during discretization by utilizing an adaptive discriminative discretization scheme, and hence greatly improves the discrimination power of classifiers. The proposed RNB+, i.e., regularized naive Bayes utilizing the proposed discretization framework, is systematically evaluated on a wide range of machine-learning datasets. It significantly and consistently outperforms state-of-the-art NB classifiers.
\end{abstract}


\end{frontmatter}

\section{Introduction}
\label{sec:intro}

Naive Bayes (NB) has been widely used in many machine-learning tasks because of its simplicity and efficiency~\citep{ren2022framework,kishwar2023fake,gonccales2022empirical,lavanya2021effective,zhang2021attribute,shaban2021accurate,geng2019model,qorib2023covid}. It could well handle different data types such as numerical and categorical ones. Naive Bayes assumes that features are independent conditioned on the classification variable, while such an assumption often does not hold~\citep{RNB2020shihe,ren2015chi}, which may degrade the classification performance. Numerous improved NB classifiers have been developed to alleviate this problem, which can be broadly divided into five categories: structure extension~\citep{wu2016sode, jiang2016structure,geng2019model}, instance selection~\citep{quinlan2014c4,wang2015adapting}, instance weighting~\citep{jiang2012discriminatively,xu2019attribute}, attribute selection~\citep{ZHANG20093218,tang2016toward, jiang2019wrapper,chen2020novel} and attribute weighting~\citep{zhang2020class, jiang2018correlation, ruan2020text, zaidi2013alleviating, jiang2019class, yu2020correlation,RNB2020shihe}. 

Among these, attribute weighting NB classifiers comparably perform better~\citep{jiang2016deep, jiang2018correlation, ruan2020text, zaidi2013alleviating, jiang2019class, yu2020correlation,RNB2020shihe}. In WANBIA, attributes are weighted differently according to the feature importance using the classification feedback~\citep{zaidi2013alleviating}. In class-specific attribute weighted naive Bayes (CAWNB), different weights are assigned to the attributes of different classes to enhance the discrimination power~\citep{jiang2019class}. Most recently, regularized naive Bayes has been developed to improve the classification performance by balancing the generalization capability and discrimination power of the classifier automatically through a gradient descent optimization using the classification feedback~\citep{RNB2020shihe}. These methods enhance the discrimination power of NB classifiers by feature weighting, but overlook the issues on data discretization.

The goal of data discretization is to find a set of cut points to optimally discretize numerical attributes, reducing the inconsistency rate while preserving the discriminant information~\citep{sharmin2019simultaneous}. However, improving generalization ability by reducing the inconsistency rate and maintaining the discrimination power are two opposite goals. On the one hand, by grouping similar values into one interval, more samples can be used to better estimate the distribution of the interval, leading to better generalization abilities, but at the cost of losing discriminant information. On the other hand, without data discretization, the discriminative ability is retained to the greatest extent, but the generalization ability is poor. A well-designed data discretization method should balance the trade-off between preserving discriminative ability and improving generalization ability.

In literature, many discretizers follow this design principle~\citep{kurgan2004caim, fayyad1993multi, tsai2008discretization}. In CAIM, a greedy algorithm is utilized to approximately find the global optimum by simultaneously minimizing the number of intervals and maximizing the class-attribute interdependence, but CAIM does not guarantee to find the global optimum~\citep{kurgan2004caim}. In MDLP, an entropy-based discretization criterion is utilized to select the cut points by maximizing the entropy of the data, which splits the attribute into intervals in a top-down manner~\citep{fayyad1993multi}. To avoid excessive splitting, a stopping criterion is defined. But this stopping criterion often leads to an early stop in the splitting process, very few discretization intervals and hence a significant information loss. Despite all these problems, it is surprising that MDLP is often used in advanced naive Bayes classifiers and yields satisfactory performance~\citep{zaidi2013alleviating,jiang2019class,RNB2020shihe,zhang2021attribute}. 

The aforementioned discretization methods are often known as supervised discretization methods~\citep{kurgan2004caim, fayyad1993multi, tsai2008discretization}, where the class information is utilized to guide the discretization process. In literature, unsupervised methods such as equal-width discretization~\citep{kurgan2004caim} and equal-frequency discretization~\citep{kurgan2004caim} are also used, which do not require the class information. The collected data are often unlabeled and labeling data is often expensive because it needs the expert knowledge~\citep{liang2019exploring}. Hence, there is often a huge amount of unlabeled data, whereas only a small portion is labeled. In this case, a semi-supervised discretization scheme is preferred to utilize both labeled and unlabeled data.

In this paper, we propose a semi-supervised adaptive discriminative discretization (SADD) to address the problem of previous methods, targeting at balancing the discrimination power and generalization ability of NB classifiers. In recent years, semi-supervised methods have been successfully applied in machine learning tasks to improve the generalization ability of models on unseen data and avoid the overfitting problem~\citep{karimi2022semiaco,hu2022multi,lai2022adaptive,lai2022semi,liang2019exploring}. In the proposed semi-supervised discretization method, unlabeled data is first assigned a pseudo label by using a simple classification model such as k-Nearest Neighbors (k-NN) classifier~\citep{hu2022multi,liang2019exploring}. Then, the pseudo-labeled data is integrated with the labeled data to provide more discriminant information for the discretization method. With the help of pseudo-labeled data, the intrinsic data structure could be better discovered in which the discriminative ability and generalization ability of subsequently trained classifiers can be greatly enhanced.

After pseudo-labeling, an adaptive discriminative discretization scheme is proposed in this paper. The proposed semi-supervised framework could better discover the intrinsic data properties, so that the data distribution could be better estimated. Data is often discretized to improve the generalization ability by grouping similar values into one interval, but too few intervals will result in a significant information loss and too few samples in the interval will result in poor generalization performance. The proposed SADD explicitly addresses the problem of early stop in MDLP~\citep{fayyad1993multi} by using an adaptive discriminative discretization scheme, and hence resolves the issue of significant discriminant information loss in MDLP. As a result, each interval has a sufficient number of samples to reliably estimate the likelihood probabilities in naive Bayes so that the naive Bayes can generalize well on unseen data, and a sufficient number of intervals are retained to maintain the discriminant information. In another word, the proposed SADD well balances the discrimination power and generalization ability of the data discretizer.

The proposed SADD is integrated with the recent development of NB classifier, regularized naive Bayes (RNB)~\citep{RNB2020shihe}, and the integrated method is named as RNB+. Compared to RNB, the proposed RNB+ well addresses the early stopping problem in data discretization of RNB and preserves the discriminant information of the data, and hence better balances the discrimination power and generalization ability. The proposed methods are evaluated on a wide range of machine-learning datasets for various applications. Experimental results show that the proposed SADD significantly outperforms the widely used discretization methods~\citep{fayyad1993multi,kurgan2004caim,tsai2008discretization} and the proposed RNB+ significantly outperforms the state-of-the-art NB classifiers~\citep{zaidi2013alleviating,jiang2019class,RNB2020shihe,zhang2021attribute}.

\section{Related Works}
\label{sec:relat_work}
We will first review various improvements of naive Bayes classifiers, especially attribute-weighting NB classifiers. Then we will review the advancements in data discretization, especially those applied to NB classifiers.

\subsection{Naive Bayes Classifiers}
Naive Bayes classifiers have been widely used in many applications~\citep{ren2022framework,kishwar2023fake,gonccales2022empirical,lavanya2021effective,zhang2021attribute,shaban2021accurate,geng2019model,qorib2023covid}. To improve the classification performance, many advanced NB classifiers have been developed, which can be divided into five categories. 1) Structure extension~\citep{wu2016sode, jiang2016structure}, by extending the structure of NB to represent the attribute dependencies, in order to alleviate the independence assumption. 2) Instance selection~\citep{quinlan2014c4,wang2015adapting}, by building a local NB model on a subset of the training instances, to reduce the effect of noisy instances. 3) Instance weighting~\citep{jiang2012discriminatively,xu2019attribute}, which enhances the discrimination power of NB classifiers by assigning a different weight to each instance. 4) Attribute selection~\citep{tang2016toward, jiang2019wrapper,chen2020novel}, to remove the redundant or irrelevant attributes and select the most informative attributes to generate the classification model. 5) Attribute weighting~\citep{jiang2016deep, jiang2018correlation, ruan2020text, zaidi2013alleviating, jiang2019class, yu2020correlation,RNB2020shihe}, by weighting the attributes differently so that the discriminative attribute has a larger weight and hence the discrimination power is increased.

Among these methods, attribute weighting methods have achieved better performance~\citep{jiang2016deep, jiang2018correlation, ruan2020text, zaidi2013alleviating, jiang2019class, yu2020correlation,RNB2020shihe}, which can be further divided into filter-based~\citep{lee2011calculating,jiang2016deep, jiang2018correlation, ruan2020text} and wrapper-based methods~\citep{zaidi2013alleviating, jiang2019class, yu2020correlation,RNB2020shihe}. The former determines the attribute weights by measuring the relationship between attributes and class variables in terms of mutual information~\citep{jiang2018correlation}, KL divergence~\citep{lee2011calculating} or correlation~\citep{jiang2018correlation}. The latter optimizes the attribute weights iteratively by using the classification feedback, which often achieves better classification performance at a higher computational cost. \cite{zaidi2013alleviating} optimized attribute weights by minimizing the mean squared error between the estimated posterior probabilities and the ones derived from ground-truth labels. Recently, \cite{jiang2019class} developed a class-specific attribute-weighting NB model, in which different weights are assigned to the attributes of different classes to enhance the discriminative ability~\citep{jiang2019class}. Very recently, \cite{RNB2020shihe} developed the regularized NB, which optimally balances the discrimination power and generalization ability of the classifier.  

\subsection{Data Discretization Methods}
In naive Bayes classifiers and many other classifiers~\citep{jiang2016deep, jiang2018correlation, ruan2020text, zaidi2013alleviating, jiang2019class, yu2020correlation,RNB2020shihe,zhang2021attribute}, numerical attributes are often discretized to group similar values into one bin, to address the problem that numerical attributes often have lots of noisy samples resulting in poor generalization capabilities. 
Existing discretization methods can be broadly divided into unsupervised, semi-supervised and supervised methods, depending on whether the class information is used~\citep{ramirez2016data}. Unsupervised discretization methods include equal-frequency, equal-width discretization~\citep{kurgan2004caim}, {proportional k-interval discretization (PKID)~\citep{yang2001proportional} and fixed frequency discretization (FFD)~\citep{yang2009discretization}}. Equal-frequency discretization divides the attribute set into intervals with the same number of instances, while the equal-width discretization divides the attribute set into intervals with the same length~\citep{kurgan2004caim}. {PKID adjusts the number and size of intervals proportional to the number of training instances~\citep{yang2001proportional}. FFD divides the data into intervals with a pre-defined frequency~\cite{yang2009discretization}} Supervised discretization methods include MDLP~\citep{fayyad1993multi}, other information-based algorithms~\citep{dougherty1995supervised,kurgan2004caim}, and statistical algorithms like ChiMerge \citep{kerber1992chimerge}, class-attribute interdependence maximization (CAIM)~\citep{kurgan2004caim} and class-attribute contingency coefficients (CACC)~\citep{tsai2008discretization}. Semi-supervised discretization methods are comparatively less studied. \cite{bondu2010non} developed a semi-supervised framework based on MODL to exploit a mixture of labeled and unlabeled data. In this paper, we mainly review CAIM~\citep{kurgan2004caim} and MDLP~\citep{fayyad1993multi} in detail as they follow closely to the design concept of balancing the discrimination power and generalization ability.
In CAIM, the boundary point is selected with the maximal interdependence in a top-down manner by using a quanta matrix~\citep{kurgan2004caim}, but the number of generated intervals is too close to the number of classes. To address this problem, \cite{tsai2008discretization} developed a discretization method based on class-attribute contingency coefficients to prevent the information loss.

Many state-of-the-art NB classifiers such as WANBIA~\citep{zaidi2013alleviating}, CAWNB~\citep{jiang2019class}, AIWNB~\citep{zhang2021attribute} and RNB~\citep{RNB2020shihe} utilize the MDLP criterion~\citep{fayyad1993multi} to discretize numerical attributes in a top-down manner. In MDLP, for each interval, a cut point with the maximum entropy amongst all candidates is selected to split the interval into two, towards the goal of retaining the maximum amount of discriminant information~\citep{fayyad1993multi}. To avoid the poor generalization ability caused by excessive splitting, a stop criterion is designed based on the concept of information encoding in a communication channel~\citep{fayyad1993multi}. MDLP often leads to a good classification performance as it balances the discrimination power and generalization ability. However, as shown later in the next section, the early stopping problem during splitting in MDLP often leads to a huge information loss and hence degrades the performance of subsequent classifiers.

\section{Proposed Method}
\label{sec:format}
\subsection{Problem Analysis of MDLP}
Data discretization is crucial to the classification performance. As an entropy-based discretization method with strong theoretical background~\citep{peng2005feature, ren2014optimizing, ren2015learning,gao2018feature,sharmin2019simultaneous,flores2019supervised}, MDLP~\citep{fayyad1993multi} has been widely used in many state-of-the-art attribute weighting NB classifiers~\citep{zaidi2013alleviating,jiang2019class,RNB2020shihe,zhang2021attribute}. It splits the dynamic range in a top-down manner, i.e., for each attribute, a cut point retaining the maximum amount of discriminant information is selected to divide the current set into two. The discriminant information is measured by the information gain of a cut point $d$ for a given attribute, dividing the current example set $\mathcal{S}$ into two subsets $\mathcal{S}_1$ and $\mathcal{S}_2$. The information gain $G(\mathcal{S}, d)$ is defined as,
\begin{equation}
G(\mathcal{S}, d) = E(\mathcal{S}) -\frac{|\mathcal{S}_1|}{|\mathcal{S}|}E(\mathcal{S}_1) - \frac{|\mathcal{S}_2|}{|\mathcal{S}|}E(\mathcal{S}_2),\label{IG}
\end{equation}
where $E(\mathcal{S})$ is the class entropy as defined below, 
\begin{equation}
E(\mathcal{S}) = -\sum_{i=1}^k P(C_i,\mathcal{S})log(P(C_i,\mathcal{S})).\label{entropy}
\end{equation}
$P(C_i, \mathcal{S})$ is the prior probability of class $C_i$ in $\mathcal{S}$, and $k$ is the number of classes. The binary split in MDLP is applied recursively if Eqn.~\eqref{mdlp} holds, and stops otherwise. Intuitively, the stop criterion will prevent excessively splitting the attribute into too many small intervals with too few samples so that the likelihood probabilities can not be reliably estimated.  
\begin{equation}
G(\mathcal{S}, d) > \theta, \label{mdlp}
\end{equation}
where $\theta$ is the adaptive threshold for discretization,
\begin{equation}
\theta = \frac{log_2(N-1)}{N}+\frac{\Delta(\mathcal{S})}{N}, \label{theta}
\end{equation}
where $N$ is the number of samples in $\mathcal{S}$. $\Delta(\mathcal{S})$ is defined as, 
\begin{equation}
\Delta(\mathcal{S}) = log_2(3^k-2) - [kE(\mathcal{S}) - k_1E(\mathcal{S}_1) - k_2E(\mathcal{S}_2)], 
\end{equation}
where $k_1$ and $k_2$ are the number of classes in $\mathcal{S}_1$ and $\mathcal{S}_2$, respectively. Empirical study shows that the threshold $\theta$ is often too large compared to the information gain $G(\mathcal{S},d)$ so that the top-down splitting often stops at an early stage. As a result, a huge amount of discriminant information is lost. 

To explore the root cause of this early stop, we take a close examination of the adaptive threshold $\theta$ defined in Eqn.~\eqref{theta}. Empirical study shows that the first term $\frac{log_2(N-1)}{N}$ dominates the adaptive threshold $\theta$, while the second term $\frac{\Delta(\mathcal{S})}{N}$ is a relatively small positive value. We further analyze $\frac{log_2(N-1)}{N}$ by plotting it against $N$ as shown in Fig.~\ref{curve2}. 

Apparently when $N$ is large, $\frac{log_2(N-1)}{N}$ is relatively small and the data could be easily split into intervals in a top-down manner. As the split continues, the number of samples $N$ in the interval becomes smaller, leading to a large $\frac{log_2(N-1)}{N}$, and hence it is more difficult to split. This decision criterion follows the design principle of balancing the discriminative ability and generalization ability, and works well when $N$ is large. However, for small $N$, $\frac{log_2(N-1)}{N}$ is relatively large and hence many attributes with a small number of samples may not split at all at the very beginning. In this case, the attribute is discretized into one bin only, and the discriminant information residing in the attribute is totally lost.

\subsection{Overview of Proposed SADD Framework for Regularized Naive Bayes}
As shown in the previous subsection, the early stopping problem in MDLP may result in a significant loss of discriminant information during discretization. In a broader sense, data discretization helps to improve the generalization ability of naive Bayes classifiers by grouping similar values into one bin, whereas excessive grouping (such as an early stop in MDLP) will result in a significant information loss. It is hence crucial for a data discretizer to balance the generalization ability and discrimination power. Furthermore, supervised discretization methods such as MDLP~\citep{fayyad1993multi} and CAIM~\citep{kurgan2004caim} are not capable of handling unlabeled data without any adaptation, and hence the information residing in unlabeled data can't be fully exploited by those supervised discretization methods. Thus, a semi-supervised discretization method exploiting both labeled and unlabeled data is needed.

\begin{figure}[!ht]
	\centering
	\includegraphics[width=1\textwidth]{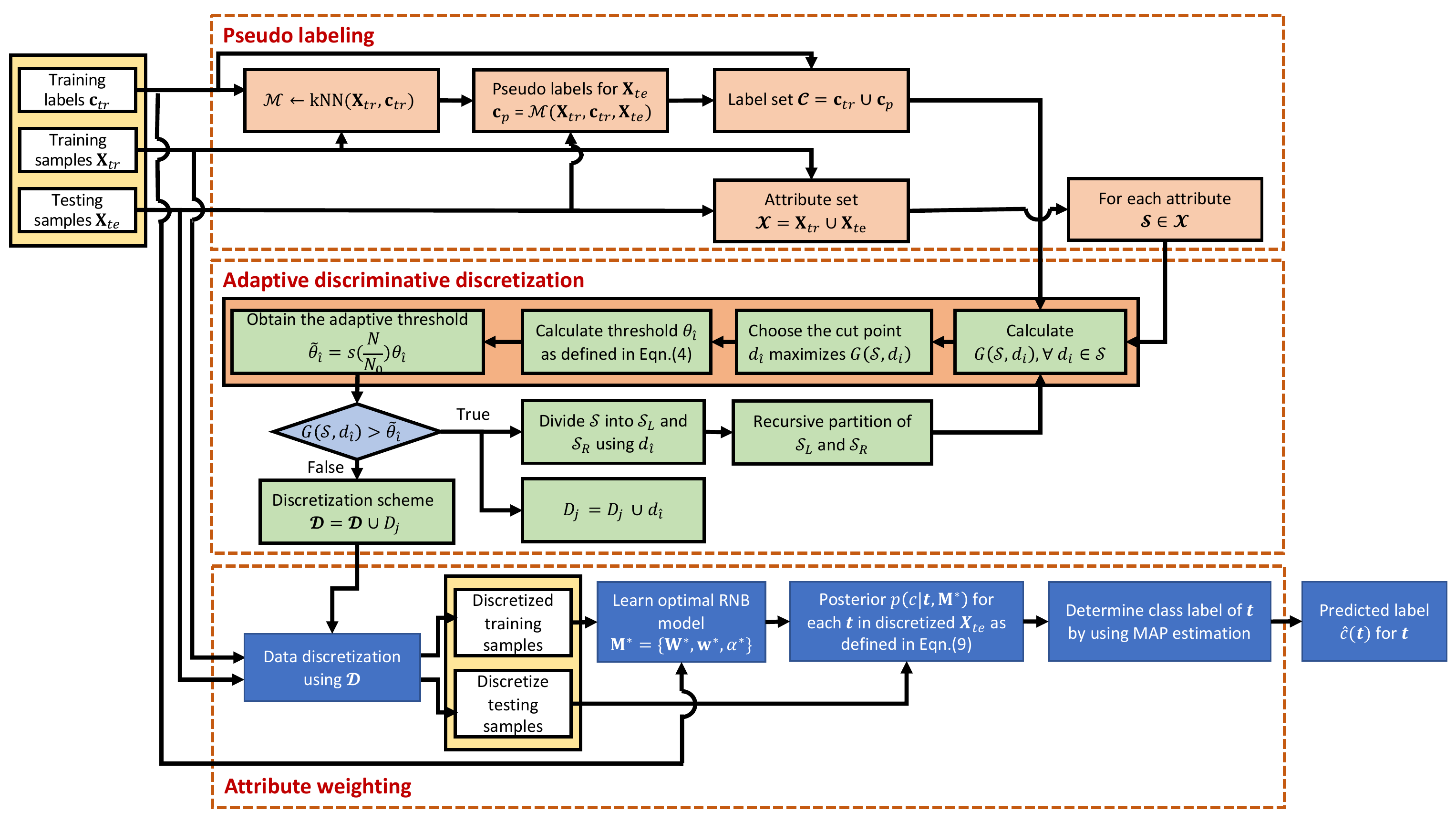}
	\caption{The proposed SADD for regularized naive Bayes. Firstly, the pseudo labels for unlabeled testing samples are generated by using k-NN. Then, both training data and testing data are used to derive the discretization scheme balancing the generalization capability and discrimination power. The derived SADD scheme is then applied to transform the numerical attributes into discrete ones, improving the generalization ability while attaining the discrimination power of the classifier. Finally, the regularized naive Bayes is integrated with the proposed SADD to derive the optimal attribute weights and determine the class labels by using MAP estimation.}
	\label{frame}
\end{figure}

To address these problems, we propose a Semi-supervised Adaptive Discriminative Discretization (SADD) method for regularized naive Bayes (The integrated method is also known as RNB+), as shown in Fig.~\ref{frame}. 1) First of all, a semi-supervised technique is designed to generate the pseudo labels for the unlabeled data, so that the intrinsic data properties in both labeled and unlabeled data could be exploited to better estimate the data statistics. In this paper, the k-NN classifier is applied to generate the pseudo labels. 2) Secondly, an adaptive discriminative discretization scheme is designed to discretize the attribute set, where a new adaptive thresholding strategy is designed to balance the number of intervals required to retain the sufficient discrimination power and the number of samples in an interval to retain the sufficient generalization ability. 3) We consider the trade-off between the generalization ability and discrimination power not only in data discretization, but also in classifier design such as feature weighting. Following this design principle, the proposed SADD is integrated with regularized naive Bayes, in which the attributes are weighted automatically balancing the discrimination power and generalization ability of the classifier.

\subsection{Proposed Semi-supervised Adaptive Discriminative Discretization}
\label{sec:SADD}
\subsubsection{Pseudo-labeling}
Semi-supervised techniques have proven to be a powerful paradigm for utilizing unlabeled data to improve the generalization ability of learning models relying solely on labeled data~\citep{karimi2022semiaco,hu2022multi,lai2022adaptive,lai2022semi}. Among various semi-supervised methods, pseudo-labeling techniques are effective to tackle the problem, which can be easily integrated with traditional supervised classification algorithms~\citep{liang2019exploring}. More specifically, let $\bm{X}_{l}$ be the labeled data with class variables $\bm{c}_{l}$ and $\bm{X}_{u}$ be the unlabeled data, the pseudo labels $\bm{c}_{p}$ for $\bm{X}_{u}$ can be derived by,
\begin{equation}
\bm{c}_{p} = \mathcal{M}(\bm{X}_{l}, \bm{c}_{l}, \bm{X}_{u}),
\label{pseudo}
\end{equation}
where $\mathcal{M}$ represents a pseudo-labeling algorithm. There are two main approaches to generate the pseudo labels: single-classifier and multi-classifier methods. In a single-classifier model, the pseudo label for each unlabeled instance is derived by using only one classification model. In contrast, multi-classifier model utilizes the majority voting rule to decide the pseudo label by using several classifiers. To keep the simplicity and effectiveness of the proposed framework, the k-nearest neighbors (k-NN) classifier is applied to generate the pseudo labels for unlabeled data. Then, the pseudo-labeled data and labeled data are combined to better discover the intrinsic data properties and better estimate the data statistics.

\subsubsection{Adaptive Discriminative Discretization}
To address the early stopping problem in previous discretization methods~\citep{fayyad1993multi}, we propose an adaptive discriminative discretization. More specifically, we aim to lower the adaptive threshold used in Eqn.~\eqref{mdlp}, especially for small datasets with relatively small $N$, in order to prevent the early stop and the significant loss of discriminant information during discretization. On the other hand, the new threshold $\tilde{\theta}$ can not be too small. If $\tilde{\theta}$ is approaching zero, each distinct value will become a separate interval, which results in no information loss, but may lead to a poor estimation of data distribution due to insufficient samples in each small interval. Bearing all these in mind, we propose the new threshold $\tilde{\theta}$ as defined below:
\begin{equation}
\tilde{\theta} = s\left(\frac{N}{N_0}\right) \theta,
\label{thresh}
\end{equation}
where $s(x) = 1/(1+e^{-x})$ is the sigmoid function and $N_0$ is a constant. As $\frac{N}{N_0}$ is non-negative, it is easy to show that $ s(\frac{N}{N_0}) \in (0.5,1)$. $N_0$ is used to judge whether there are sufficient samples in the interval. If $N \gg N_0$, i.e., there are sufficient samples, $s(\frac{N}{N_0}) \approx 1$. In this case, although $s(\frac{N}{N_0})$ is relatively large, $\theta$ is relatively small, and $\tilde{\theta}$ is small enough so that the top-down split could continue, and the resulting intervals will have sufficient samples to reliably estimate the likelihood probabilities. If $N \approx N_0$, $s(\frac{N}{N_0}) \approx 0.73$, i.e., a significantly lower threshold $\tilde{\theta}$ will be used in the top-down discretization compared to the threshold $\theta$ defined in Eqn.~\eqref{theta}. Consequently, it will encourage further splitting of the interval and hence retain more discriminant information. To prevent excessive splitting, the proposed method has a safeguard mechanism. More specifically, when $N\ll N_0$, $s(\frac{N}{N_0}) \approx 0.5$, i.e., the maximum reduction of the threshold $\tilde{\theta}$ from $\theta$ is 50\%. 
\begin{figure}[ht]
	\centering
	\includegraphics[width=0.7\textwidth]{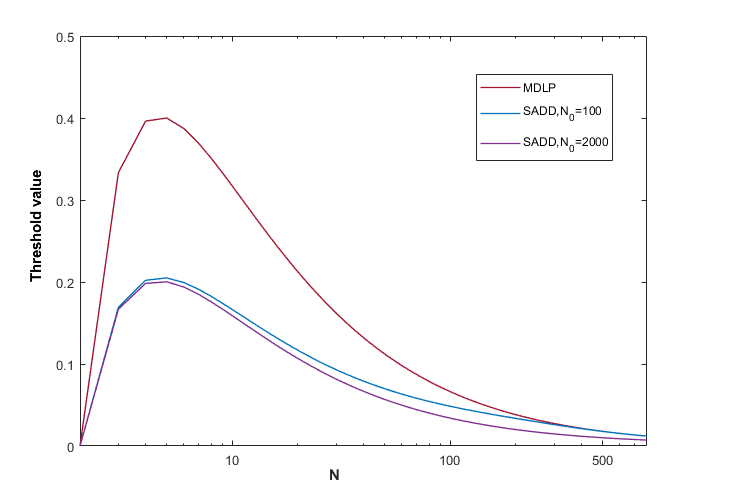}
	\caption{Plot of $\frac{log_2(N-1)}{N}$ and its adaptive version $s(\frac{N}{N_0})\frac{log_2(N-1)}{N}$ with different values of $N_0$. When $N$ is small, the new adaptive threshold is about half of $\frac{log_2(N-1)}{N}$, which prevents a possible early stop in the top-down split.} 
	\label{curve2}
\end{figure}

To further illustrate the effect of the adaptive threshold in the proposed method, we plot the value of the adaptive version of $\frac{log_2(N-1)}{N}$ after multiplying it with $s(\frac{N}{N_0})$ for different $N_0$, as shown in Fig.~\ref{curve2}. With the small number of samples in an interval, the threshold is greatly decreased from about 0.4 to 0.2 to encourage further splitting, especially for small datasets. When $N$ is large, the new adaptive threshold is smaller than the original one, but very close to it, to encourage the split. In addition, the proposed method is insensitive to the choice of $N_0$. As shown in Fig.~\ref{curve2}, the new thresholds for $N_0=100$ and $N_0 = 2000$ are quite close when $N$ is small. For large $N$, the difference does not matter so much as the set will be split into smaller ones anyway. In summary, the proposed SADD could effectively prevent the early stop during data discretization, as well as the excessive split. The proposed SADD algorithm is summarized in Algorithm~\ref{alg}.
\begin{algorithm}[!hp]
	\footnotesize
	\renewcommand{\algorithmicrequire}{\textbf{Input:} }
	\renewcommand{\algorithmicensure}{\textbf{Output:}}
	\caption{The proposed SADD algorithm}
	\label{alg}
	\begin{algorithmic}[1]
		\Require Training data $\bm{X}_{l}$ with class $\bm{c}_{l}$ and testing data $\bm{X}_{u}$
		\Ensure Discretization scheme $\bm{\mathcal{D}} \gets \{\mathcal{D}_1, \mathcal{D}_2,..., \mathcal{D}_m\}$  for $m$-dimensional features
		\State $\bm{c}_{p} \gets \mathcal{M}(\bm{X}_{l}, \bm{c}_{l}, \bm{X}_{u})$ \Comment{Derive pseudo labels using Eqn.~(\ref{pseudo})}
		\State $\bm{\mathcal{X}} \gets \bm{X}_{l} \cup \bm{X}_{u}$  \Comment{$\bm{\mathcal{X}}$ is the set of samples}
		\State $\bm{\mathcal{C}} \gets \bm{c}_{l} \cup \bm{c}_{p}$	\Comment{$\bm{\mathcal{C}}$ is the set of labels}
		\State $\bm{\mathcal{D}} \gets \emptyset$ \Comment{Initialize $\bm{\mathcal{D}}$ as an empty set}
		\For{each $\mathcal{X}_j \in \bm{\mathcal{X}}$}
		\State $\mathcal{D}_j \gets \emptyset$ \Comment{$\mathcal{D}_j$ is the discretization scheme for $\mathcal{X}_j$}
		\State $\mathcal{S} \gets \mathcal{X}_j$ \Comment{$\mathcal{S}$ is the set of samples to be discretized}
		\Procedure{Partition}{$\mathcal{S},\bm{\mathcal{C}}$} \Comment{Procedure to partition $\mathcal{S}$ using $\bm{\mathcal{C}}$}
		\If{$|\mathcal{S}|$ ==1} \Comment{If there is only one sample in $\mathcal{S}$, return}
		\State \Return
		\EndIf
		\State Calculate $G(\mathcal{S},d_i), \forall d_i\in \mathcal{S}$, as defined in Eqn.~(\ref{IG}).
		\State Choose the cut point $d_{\hat{i}}$, $\hat{i} = \mathop{\arg\max}_{i}{G(\mathcal{S},d_i)}, \forall d_i\in \mathcal{S}$.
		\State Calculate the threshold $\theta_{\hat{i}}$ using Eqn.~(\ref{theta}) for the cut point $d_{\hat{i}}$.
		\State Calculate the new adaptive threshold $\tilde{\theta}_{\hat{i}} = s\left(\frac{N}{N_0}\right) \theta_{\hat{i}}$.
		\If{$G(\mathcal{S},d_{\hat{i}}) > \tilde{\theta}_{\hat{i}}$} \Comment{The SADD stop criterion}
		\State $\mathcal{D}_j \gets \mathcal{D}_j \cup d_{\hat{i}}$ \Comment{Insert $d_{\hat{i}}$ into discretization scheme $\mathcal{D}_j$}
		\State $\mathcal{S}_L \gets \mathcal{S}<d_{\hat{i}}$  \Comment{Divide $\mathcal{S}$ into the set smaller than $d_{\hat{i}}$}
		\State $\mathcal{S}_R \gets \mathcal{S} \geq d_{\hat{i}}$ \Comment{Divide $\mathcal{S}$ into the set not smaller than $d_{\hat{i}}$}
		\State \Call{Partition}{$\mathcal{S}_L,\bm{\mathcal{C}}$} \Comment{Recursively partition $\mathcal{S}_L$ using $\bm{\mathcal{C}}$}
		\State \Call{Partition}{$\mathcal{S}_R,\bm{\mathcal{C}}$} \Comment{Recursively partition $\mathcal{S}_R$ using $\bm{\mathcal{C}}$}
		\EndIf
		\EndProcedure
		\State $\bm{\mathcal{D}} \gets \bm{\mathcal{D}} \cup \mathcal{D}_j$ \Comment{Add the discretization scheme $\mathcal{D}_j$ into $\bm{\mathcal{D}}$}
		\EndFor
		\State \Return $\bm{\mathcal{D}}$
	\end{algorithmic}
\end{algorithm}

In the first step of Algorithm~\ref{alg}, pseudo labeling, a k-NN classification model $\mathcal{M}$ is generated by using the labeled training data and used to derive the pseudo labels $\bm{c}_{p}$ for unlabeled testing data using Eqn.~(\ref{pseudo}). Then, the attribute set $\bm{\mathcal{X}}$ and label set $\bm{\mathcal{C}}$ are derived by combining the labeled training data with pseudo-labeled testing data. Then the attributes $\mathcal{X}_j\in \bm{\mathcal{X}}$ are discretized one at a time. The procedure PARTITION is used to find the optimal cut point $d_{\hat{i}}$ to divide the current sample set $\mathcal{S}$ into two sets $\mathcal{S}_L$ and $\mathcal{S}_R$, where $\hat{i} = \mathop{\arg\max}_{i}{G(\mathcal{S},d_i)}, \forall d_i\in \mathcal{S}$ and $G(\mathcal{S},d_i)$ is the information gain defined in Eqn.~(\ref{IG}). Then, the PARTITION procedure is recursively applied on $\mathcal{S}_L$ and $\mathcal{S}_R$ to find the optimal cut point to further discretize the attribute. The recursive partition continues as long as the following condition holds:
\begin{equation}
	G(\mathcal{S}, d_{\hat{i}}) > \tilde{\theta}_{\hat{i}},
	\label{eqn:G_new_T}
\end{equation}
where $\tilde{\theta}_{\hat{i}} = s\left(\frac{N}{N_0}\right) \theta_{\hat{i}}$ is the newly defined adaptive threshold, $\theta_{\hat{i}}$ is the threshold defined in Eqn.~(\ref{theta}) for the cut point $d_{\hat{i}}$ and $s(\cdot)$ is the sigmoid function. The discretization scheme $\mathcal{D}_j$ for attribute $\mathcal{X}_j$ is updated as,
\begin{equation}
	\mathcal{D}_j \gets \mathcal{D}_j \cup d_{\hat{i}}.
\end{equation}
{For each attribute, the proposed SADD utilizes a greedy hierarchical splitting algorithm to generate a tree-like discretization scheme, as summarized in Algo.~\ref{alg}. It can be shown that the time complexity is $O(n\log{n})$ for each attribute, where $n$ is the number of samples. The total time complexity for $m$ attributes is hence $O(mn\log{n})$.}

\subsection{Discussion and Analysis}
\begin{table}[!ht]
\caption{The comparisons of the number of intervals (Num.) and the mutual information (MI), after data discretization by the proposed SADD and MDLP~\citep{fayyad1993multi} on all numerical attributes of the ``Vowel" dataset.}
	\centering
\begin{tabular}{|c|cc|cc|}
\hline
     & \multicolumn{2}{c|}{Proposed SADD} & \multicolumn{2}{c|}{MDLP}          \\ \hline
     & \multicolumn{1}{c|}{MI}     & Num. & \multicolumn{1}{c|}{MI}     & Num. \\ \hline
$A_1$  & \multicolumn{1}{c|}{1.0921} & 18   & \multicolumn{1}{c|}{0.9596} & 8    \\ \hline
$A_2$  & \multicolumn{1}{c|}{1.2180} & 19   & \multicolumn{1}{c|}{1.0875} & 9    \\ \hline
$A_3$  & \multicolumn{1}{c|}{0.2998} & 8    & \multicolumn{1}{c|}{0.1198} & 3    \\ \hline
$A_4$  & \multicolumn{1}{c|}{0.4488} & 7    & \multicolumn{1}{c|}{0.3545} & 4    \\ \hline
$A_5$  & \multicolumn{1}{c|}{0.5909} & 14   & \multicolumn{1}{c|}{0.4104} & 4    \\ \hline
$A_6$  & \multicolumn{1}{c|}{0.4347} & 11   & \multicolumn{1}{c|}{0.2759} & 4    \\ \hline
$A_7$  & \multicolumn{1}{c|}{0.3287} & 7    & \multicolumn{1}{c|}{0.2146} & 3    \\ \hline
$A_8$  & \multicolumn{1}{c|}{0.2447} & 7    & \multicolumn{1}{c|}{0.1685} & 3    \\ \hline
$A_9$  & \multicolumn{1}{c|}{0.3009} & 7    & \multicolumn{1}{c|}{0.1796} & 3    \\ \hline
$A_{10}$ & \multicolumn{1}{c|}{0.0527} & 2    & \multicolumn{1}{c|}{0}      & 1    \\ \hline
AVG  & \multicolumn{1}{c|}{0.5011} & 10   & \multicolumn{1}{c|}{0.3770} & 4.2  \\ \hline
\end{tabular}
\label{case}
\end{table}

As discussed early, the proposed SADD could effectively prevent the information loss of MDLP. To further analyze this, a case study is presented in Table~\ref{case}, which shows the number of intervals (Num.) and the mutual information (MI) on the ``Vowel" dataset after discretization by MDLP~\citep{fayyad1993multi} and the proposed SADD, respectively. The dataset contains 10 numerical attributes and 11 classes. Most numerical attributes are discretized into 3-4 intervals by MDLP, which can not effectively differentiate 11 classes. After applying the proposed SADD, more intervals could be obtained and hence less discriminant information is lost, as shown in Table~\ref{case}. The proposed SADD can effectively prevent information loss and generate a discretization scheme with a better trade-off between the number of intervals and the number of samples in the intervals. On the one hand, more intervals will retain more discriminant information, but lead to a poor generalization ability as there are too few samples in an interval to reliably estimate the data distribution. On the other hand, too few intervals may result in a huge discriminant information loss, as shown in the early stopping case of MDLP. 

\subsection{Proposed RNB+}
MDLP has been widely used in many state-of-the-art naive Bayes classifiers, e.g., AIWNB~\citep{zhang2021attribute}, WANBIA~\citep{zaidi2013alleviating}, CAWNB~\citep{jiang2019class} and RNB~\citep{RNB2020shihe}. As shown previously, MDLP may result in a huge information loss and hence the SADD is proposed to address this problem. In this section, we describe how to integrate the proposed SADD with RNB to boost the classification performance of NB classifiers. The integrated method is named as RNB+. We first discretize the data using the proposed SADD so that the data distribution could be better estimated, and then use RNB as the classifier. 

We have shown that the proposed SADD could well balance the generalization ability and discrimination power during data discretization. Now we show how the proposed RNB+ achieves a better trade-off during attribute weighting. To alleviate the conditional independence assumption of NB classifiers, attribute weighting techniques have been widely used in NB classifiers and achieved remarkable performance~\citep{zaidi2013alleviating,jiang2019class,RNB2020shihe,zhang2021attribute}. In WANBIA, the same weight is assigned to the attributes in different classes~\citep{zaidi2013alleviating}, while in CAWNB, a class-specific weight is assigned to each attribute to capture more data characteristics~\citep{jiang2019class}. But the model complexity increases with more attribute weights, and CAWNB hence may overfit to the data, especially for small datasets. To alleviate this problem, regularized naive Bayes has been recently developed, which regularizes CAWNB by adding a simpler model, i.e., WANBIA, to penalize the model complexity~\citep{RNB2020shihe}. 

More specifically, in RNB, the target is to find the optimal model parameters $\bm{M}$ =$ \{\bm{W}, \bm{w}, \alpha\}$ to minimize the difference between the posterior derived from the ground-truth label and the estimated posterior for a given instance $\bm{x}$,
\begin{equation}
P(c|\bm{x},\bm{M}) = \alpha P_D(c|\bm{x},\bm{W}) + (1-\alpha) P_I(c|\bm{x},\bm{w}),
\label{regular-nb}
\end{equation}
where $P_D(c|\bm{x},\bm{W})$ is the posterior where attributes are weighted on a class-specific basis and $\bm{W}$ is the weight matrix. $P_I(c|\bm{x},\bm{w})$ is the posterior where attributes are weighted the same for all classes and $\bm{w}$ is the weight vector. $P_D(c|\bm{x},\bm{W})$ is a more complex model that could provide more discrimination power, whereas $P_I(c|\bm{x},\bm{w})$ is a simpler model that can provide better generalization ability. In RNB, the optimal model parameters $\bm{M^*}$ are derived through a gradient-descent algorithm, and the discrimination power and generalization ability are automatically balanced by optimizing $\alpha$~\citep{RNB2020shihe}. Finally, the predicted label for each test instance $\bm{t}$ is obtained by using the MAP estimation as follows: 
\begin{equation}
\hat{c}(\bm{t}) =\mathop{\arg\max}_{c \in C} P(c|\bm{t},\bm{M}^*), \label{map}
\end{equation}
where $C$ is the set of labels for all classes.

The categorical attributes and numerical attributes are often mixed and NB classifiers can generate a probabilistic model on both data types. However, the numerical attributes often have a large number of distinct values so that the likelihood probability estimated from the frequency of instances with a particular value $x_i$ in the $j$-th attribute given the class $c$, $P(A_{j}=x_{i}|c)$, can be extremely small. The estimation of $P(A_{j}=x_{i}|c)$ may not be reliable due to very few training instances. To address this problem, discretization methods have been developed by grouping similar values into one interval and then sufficient training instances can be used to reliably estimate the likelihood probability. However, many discretization methods, e.g., MDLP~\citep{fayyad1993multi} in the state-of-the-art NB classifiers~\citep{zaidi2013alleviating,jiang2019class,zhang2021attribute,RNB2020shihe}, can't generate a proper discretization scheme and may lead to the huge information loss. Thus, RNB+ is proposed to alleviate this problem and retain the discriminative ability from discretization perspective. 
As shown later in the experiments, the proposed RNB+ significantly outperforms the state-of-the-art NB classifiers such as RNB~\citep{RNB2020shihe}, WANBIA~\citep{zaidi2013alleviating}, CAWNB~\citep{jiang2019class} and AIWNB~\citep{zhang2021attribute}.

\section{Experimental Results}
\label{sec:exp-res}
\subsection{Experimental Settings}
The experiments are divided into two parts under NB classification framework~\citep{webb2003statistical}. The proposed SADD is firstly compared with other discretization methods including four supervised discretization, MDLP~\citep{fayyad1993multi}, CAIM~\citep{kurgan2004caim}, CACC~\citep{tsai2008discretization} and ChiMerge~\citep{kurgan2004caim}, and four unsupervised discretization, Equal-Frequency~\citep{kurgan2004caim}, Equal-Width~\citep{kurgan2004caim} \rev{PKID~\citep{yang2001proportional} and FFD~\citep{yang2009discretization}}. Then, the proposed RNB+ is compared with RNB~\citep{RNB2020shihe}, WANBIA~\citep{zaidi2013alleviating}, CAWNB~\citep{jiang2019class} and AIWNB~\citep{zhang2021attribute}, which are four recent attribute-weighting NB classifiers. All the competitors are summarized in Table~\ref{comp}. \rev{The experimental results of PKID~\citep{yang2001proportional} and FFD~\citep{yang2009discretization} are obtained by using the popular data mining tool, KEEL~\citep{alcala2009keel}. The other competitors are implemented by using MATLAB.} \rev{In the proposed SADD, k-NN classifier with Euclidean distance is used in pseudo-labeling, where the optimal $k$ is tuned by using the validation set. Specifically, one out of nine folds of the training data is randomly selected as the validation set. The optimal $k$ is derived by using a grid search that produces the highest classification accuracy on the validation set.} 
\begin{table}[ht]
\centering
  \caption{Description of competitors: six popular discretization methods and four state-of-the-art NB classifiers.}
  \begin{tabular}{|p{80pt}<{\centering}|p{230pt}|}
  \hline
  \multicolumn{2}{|c|}{\textbf{Discretization methods}} \\   \hline
  MDLP    & Supervised entropy-based top-down discretization                  \\ \hline
  CAIM& Supervised statistical top-down discretization \\ \hline
  CACC&Supervised statistical top-down discretization \\ \hline
  ChiMerge&Supervised statistical bottom-up discretization \\ \hline
  Equal-W& Unsupervised top-down discretization \\ \hline
  Equal-F&Unsupervised top-down discretization \\ \hline 
  \rev{PKID}&\rev{Unsupervised top-down discretization}\\ \hline
  \rev{FFD}&\rev{Unsupervised top-down discretization}\\ \hline
  \multicolumn{2}{|c|}{\textbf{Naive Bayes methods}} \\ \hline
  WANBIA & Wrapper-based class-independent attribute weighting               \\ \hline
  CAWNB&Wrapper-based class-specific attribute weighting\\ \hline
  AIWNB& Filter-based attribute and instance weighting\\ \hline
  RNB   & Wrapper-based regularized attribute weighting \\ \hline
  \end{tabular}
  \label{comp}
  \end{table}
  
The comparison experiments are conducted on a set of machine-learning datasets in various domains such as healthcare, biology, disease diagnosis and business. All the datasets are extracted from the UCI machine learning repository~\footnote{https://archive.ics.uci.edu/ml/index.php}.
Among them, 12 datasets were used in CACC~\citep{tsai2008discretization} and the rest of them are selected to enrich the comparison experiments. The number of instances is distributed between 150 and 21048 and the number of attributes is between 4 and 520. The numerical attributes and categorical attributes are mixed in the datasets. Some datasets have missing values, which are replaced by the mean of corresponding numerical attributes or mode of categorical attributes. These 31 benchmark datasets provide a comprehensive evaluation of the proposed SADD and RNB+. The datasets are summarized in Table~\ref{dataset}. 
\begin{table}[!hp]
	\centering
	\caption{\footnotesize{Most of the datasets are collected from real-world problems in various domains. The number of instances is distributed between 150 and 19020, which provides a comprehensive evaluation on different sizes of datasets. The number of attributes and classes on these datasets are significantly different. Some datasets contain missing values which present real-world difficulties when collecting the data. For the entry $u(v)$ in ``Attribute", $u$ denotes the total number of attributes and $v$ denotes the number of categorical attributes.}}
	\resizebox{.65\columnwidth}{!}{
	\begin{tabular}{ccccc}
		\toprule
		Dataset	&	Instance	&	Attribute	&	Class	&	Missing values	\\ \midrule
Iris           & 150      & 4         & 3     & N              \\
Parkinson      & 195      & 23        & 2     & N              \\
Seeds          & 210      & 7         & 3     & N              \\
Glass          & 214      & 10        & 6     & N              \\
Heart          & 270      & 13(7)     & 2     & N              \\
Ecoli          & 336      & 8         & 8     & N              \\
Bupa           & 345      & 6         & 2     & N              \\
Ionophere      & 351      & 34(2)     & 2     & N              \\
Movement       & 360      & 90        & 15    & N              \\
ILPD           & 583      & 10        & 2     & N              \\
Breast         & 699      & 9         & 2     & Y              \\
Pima           & 768      & 8         & 2     & N              \\
Vowel          & 990      & 13        & 11    & N              \\
Biodegradation & 1055     & 41        & 2     & N              \\
Mice Protein   & 1080     & 82        & 8     & Y              \\
Yeast          & 1484     & 10        & 8     & N              \\
\rev{Mfeat-fac}         & \rev{2000}    & \rev{216}        & \rev{10}     & \rev{N}  \\
Cardio         & 2126     & 23        & 10    & N              \\
Madelon        & 2600     & 500       & 2     & N              \\
Spambase       & 4601     & 57        & 2     & N              \\
Wave           & 5000     & 40        & 3     & N              \\
Wall-Following & 5456     & 24        & 4     & Y              \\
Page-Block     & 5473     & 10        & 5     & N              \\
Opdigit        & 5620     & 64        & 10    & N              \\
Satellite      & 6435     & 36        & 6     & N              \\
Wine           & 6497     & 11        & 7     & N              \\
\rev{Musk}         & \rev{6598}    & \rev{166}        & \rev{2}     & \rev{N}  \\
Anuran         & 7195     & 22        & 4     & N              \\
Pendigit       & 10992    & 16        & 10    & N              \\
Magic          & 19020    & 10        & 2     & N              \\
\rev{IndoorLoc}         & \rev{21048}    & \rev{520}        & \rev{3}    & \rev{N}              \\
		\bottomrule
	\end{tabular}
}
	\label{dataset}
\end{table}

\subsection{Comparisons to State-of-the-art Discretization Methods}
The proposed SADD is compared with MDLP~\citep{fayyad1993multi}, CAIM~\citep{kurgan2004caim}, CACC~\citep{tsai2008discretization}, ChiMerge~\citep{kerber1992chimerge}, Equal-W~\citep{kurgan2004caim}, Equal-F~\citep{kurgan2004caim}, \rev{PKID~\citep{yang2001proportional} and FFD~\citep{yang2009discretization}} based on the NB classifier~\citep{webb2003statistical}. 
Table~\ref{res} summarizes the comparisons to these discretization methods. The classification accuracy of each algorithm on each dataset is derived via stratified 10-fold cross-validation, following the same evaluation protocol used in~\citep{jiang2019class, zaidi2013alleviating, zhang2021attribute, RNB2020shihe}. The symbol $\bullet$ in Table~\ref{res} indicates that the proposed method significantly outperforms its competitors with a one-tailed t-test at the significance level of $p=0.05$. The average classification accuracies of all algorithms over all the datasets are summarized at the bottom, which can provide a straightforward comparison of different methods. The hyper-parameter $N_0$ in the proposed SADD is set to 2000 empirically. 

\begin{table}[!ht]
\caption{\scriptsize{\rev{Comparisons between the proposed SADD and other discretization methods such as MDLP~\citep{fayyad1993multi}, CAIM~\citep{kurgan2004caim}, CACC~\citep{tsai2008discretization}, ChiMerge~\citep{kerber1992chimerge}, Equal-W (EW)~\citep{kurgan2004caim}, Equal-F (EF)~\citep{kurgan2004caim}, PKID~\citep{yang2001proportional} and FFD~\citep{yang2009discretization} based on the NB classifier~\citep{webb2003statistical}. The proposed SADD achieves an average performance gain of 3.11\% compared with MDLP, and the performance gain is 2.80\% on average compared with previously best performed method, CAIM.}}}
\resizebox{1\columnwidth}{!}{
\begin{tabular}{@{}cccccccccc@{}}
\toprule
Dataset        & SADD                   & MDLP         & CAIM         & CACC                  & ChiMerge              & EW                    & EF           & PKID                  & FFD                   \\ \midrule
Iris           & \textBF{96.00±4.42}  & 92.67±6.29 & 94.00±5.54 & 93.33±6.67          & 78.67±9.80          & 94.67±6.53          & 92.67±8.14 & 91.33±9.45          & 93.33±6.67          \\
Parkinson      & \textBF{84.02±5.65}  & 79.46±4.69 & 81.44±7.29 & 82.46±6.79          & 81.02±9.18          & 79.94±6.41          & 80.35±7.25 & 77.26±9.90          & 77.26±7.84          \\
Seeds          & 90.95±6.19           & 87.14±4.29 & 86.67±5.13 & 87.62±5.30          & 80.95±4.26          & \textBF{91.43±5.13} & 88.57±5.71 & 90.00±8.64          & 87.62±8.83          \\
Glass          & \textBF{74.29±4.82} & 72.03±8.65 & 72.92±9.36 & 65.25±11.14         & 66.40±10.08         & 60.75±7.68          & 68.24±9.01 & 65.35±8.78          & 65.71±10.17         \\
Heart          & 83.70±8.64          & 83.70±8.64 & 83.33±8.16 & 80.74±6.58          & 83.70±9.40          & \textBF{84.44±7.73} & 82.96±8.31 & 82.96±4.74          & 83.70±3.78          \\
Ecoli          & \textBF{86.62±4.38}  & 83.10±4.22 & 81.46±5.58 & 82.47±5.97          & 83.38±6.43          & 85.10±4.26          & 84.28±5.64 & 81.58±7.04          & 82.45±3.05          \\
Bupa           & \textBF{65.76±10.42}  & 53.27±9.52 & 65.24±6.98 & 63.24±6.30          & 64.03±8.58          & 62.61±8.22          & 58.55±7.53 & 61.73±7.41          & 62.87±8.43          \\
Ionophere      & \textBF{90.62±5.19}  & 89.52±5.12 & 88.64±4.11 & 88.92±4.40          & 75.83±6.21          & 90.32±4.41          & 89.77±5.47 & 89.16±6.11          & 89.17±5.09          \\
Movement       & \textBF{77.77±6.72}  & 62.10±7.37 & 71.97±7.94 & 71.14±7.41          & 71.41±6.83          & 70.32±7.17          & 72.18±6.59 & 64.87±8.37          & 67.37±10.45         \\
ILPD           & 67.05±4.18          & 64.82±4.45 & 65.51±4.28 & 66.03±4.19          & 65.00±2.68          & 67.75±2.18          & 67.40±4.27 & 67.56±4.89          & \textBF{68.09±2.91} \\
Breast         & 97.42±1.41  & 97.14±1.43 & 97.28±1.50 & 97.28±1.50          & 95.85±2.67          & 97.42±1.55          & 97.42±1.67 & \textBF{97.43±1.90}          & 97.43±1.90          \\
Pima           & 76.82±4.34           & 73.69±4.70 & 74.21±5.80 & 74.21±3.92          & 72.26±5.17          & \textBF{77.21±2.80} & 74.61±3.61 & 74.82±4.37          & 75.35±4.80          \\
Vowel          & 75.76±4.93           & 59.09±4.45 & 64.34±5.27 & 60.71±6.69          & 61.11±3.23          & 67.58±4.87          & 64.24±5.15 & \textBF{89.63±1.37} & 60.00±6.69          \\
Biodegradation & \textBF{81.89±2.80}  & 80.85±2.76 & 81.70±2.48 & 81.70±2.75          & 78.67±4.12          & 80.19±2.08          & 81.04±3.19 & 80.38±3.22          & 80.00±3.10          \\
Mice Protein   & \textBF{98.06±0.77}  & 93.98±2.60 & 93.34±2.63 & 91.67±2.60          & 92.40±3.55          & 94.63±2.56          & 93.05±2.61 & 93.14±2.05          & 93.79±2.34          \\
Yeast          & \textBF{59.79±3.89}  & 57.15±3.56 & 57.49±3.54 & 56.01±2.88          & 58.23±4.60          & 58.64±4.80          & 55.68±4.93 & 54.31±2.71          & 53.57±2.81          \\
Mfeat-fac 	   & \textBF{94.80±1.95} & 93.15±2.15 & 93.70±2.02 & 93.70±2.23 & 93.15±2.08 & 93.3±2.35 & 92.70±2.50 & 92.15±2.08 &  82.79±2.42\\
Cardio         & 81.19±1.41           & 79.68±1.66 & 80.34±1.78 & 79.35±1.38          & 78.12±1.83          & 79.25±1.61          & 77.80±2.10 & 80.15±1.76          & \textBF{81.28±2.76} \\
Madelon        & \textBF{64.65±3.79}  & 61.92±3.34 & 58.69±2.85 & 50.00±0.00          & 58.96±3.98          & 50.00±0.00          & 50.00±0.00 & 55.35±2.47          & 52.92±3.62          \\
Spambase       & 90.18±1.72           & 89.63±1.45 & 89.85±1.53 & 90.00±1.63          & 89.42±1.01          & 85.53±1.97          & 89.87±1.41 & \textBF{95.60±0.85} & 89.33±1.39          \\
Wave           & \textBF{80.70±1.00}  & 80.18±1.00 & 80.60±1.30 & 80.32 ± 1.28          & 78.80±1.04          & 80.16±0.83          & 80.24±0.86 & 79.10±1.60          & 78.60±1.18          \\
Wall-Following & \textBF{90.96±0.96}  & 89.24±1.29 & 88.11±1.06 & 87.81±1.11          & 71.15±1.60          & 80.96±1.41          & 84.26±1.33 & 60.91±3.62          & 86.33±1.42          \\
Page-Block     & \textBF{93.93±1.41}  & 93.62±1.62 & 93.17±1.25 & 93.79±1.22          & 91.25±1.18          & 92.54±0.80          & 88.86±1.57 & 91.78±1.07          & 92.98±0.81          \\
Opdigit        & \textBF{92.65±0.51}  & 92.38±0.40 & 92.22±0.64 & 92.31±0.82          & 91.76±0.95          & 92.46±0.74          & 91.80±0.93 & 92.19±1.16          & 92.15±1.15          \\
Satellite      & \textBF{82.45±1.46}  & 82.14±1.40 & 82.02±1.43 & 82.08±1.40          & 79.52±1.55          & 81.18±1.16          & 81.15±1.31 & 82.10±1.43          & 82.14±1.49          \\
Wine           & 50.42±1.00           & 49.19±1.42 & 49.92±0.86 & \textBF{50.99±1.62} & 48.47±1.49          & 49.42±1.46          & 47.87±1.07 & 51.82±1.43          & \textBF{53.21±1.78} \\
Musk 		   & \textBF{92.98±0.79} & 91.76±0.93 & 85.50±1.54 & 89.83±0.75 & 81.60±2.05 & 84.18±1.85 & 83.62±1.54 & 91.13±0.78 & 61.68±1.61 \\
Anuran         & \textBF{90.62±1.09}  & 89.92±1.24 & 89.42±1.29 & 89.26±1.41          & 81.86±1.38          & 89.33±1.30          & 89.01±1.00 & 89.41±1.25          & 88.27±1.57          \\
Pendigit       & \textBF{88.43±0.61}  & 88.10±0.84 & 87.92±0.72 & 88.07±0.75          & 86.96±0.65          & 87.33±0.83          & 87.25±0.82 & 87.24±0.91          & 86.61±0.79          \\
Magic          & \textBF{78.13±0.49}  & 77.67±0.56 & 75.57±0.57 & 76.15±0.68          & 73.26±0.78          & 74.67±0.51          & 76.55±0.81 & 77.78±0.89          & 77.32±0.82          \\
IndoorLoc 	   & \textBF{65.55±0.76}  & 59.29±1.22   & 64.18±0.84  & 64.80±0.80  & 59.24±0.81   & 59.94±1.25  & 41.65±0.86  & 61.68±0.78   & 63.19±0.52  \\ \midrule

\textBF{AVG}        & 82.07 & 78.96 & 79.27 & 79.07 & 76.58 & 78.81 & 77.86 & 78.58 & 78.10      \\ \bottomrule
\end{tabular}
}
\label{res}
\end{table}

As shown in Table~\ref{res}, compared to other discretization methods, the proposed SADD achieves the highest average classification accuracy. On most of the datasets, the  proposed method performs best, and many of the performance gains are statistically significant. Compared with MDLP~\citep{fayyad1993multi}, CAIM~\citep{kurgan2004caim}, CACC~\citep{tsai2008discretization} and ChiMerge~\citep{kerber1992chimerge}, the proposed SADD obtains the performance gain of 3.11\%, 2.80\%, 3.00\% and	5.49\%  on average, respectively. \rev{Compared with the four unsupervised discretization methods, Equal-W~\citep{kurgan2004caim}, Equal-F~\citep{kurgan2004caim}, PKID~\citep{yang2001proportional} and FFD~\citep{yang2009discretization}, the proposed SADD obtains the average performance gain of 3.26\%, 4.21\%, 3.49\% and 3.97\%, respectively.} These results demonstrate the superior performance of the proposed SADD. 

Table~\ref{sum} summarizes the results for statistical significance tests. The proposed SADD achieves the best performance on most of the datasets, and the performance gains on many of them are statistically significant. Specifically, the proposed SADD outperforms MDLP~\citep{fayyad1993multi}, CAIM~\citep{kurgan2004caim}, CACC~\citep{tsai2008discretization}, ChiMerge~\citep{kerber1992chimerge}, Equal-W~\citep{kurgan2004caim}, Equal-F~\citep{kurgan2004caim}, PKID~\citep{yang2001proportional} and FFD~\citep{yang2009discretization} on 31, 31, 30, 31, 27, 30, 27 and 26 datasets respectively, among which 19, 13, 15, 25, 18, 21, 19 and 18 are statistically significant.
\begin{table}[!ht]
	\centering
	\caption{Summary of statistical significance tests on different discretization methods. For each entry $u(v)$, $u$ is the number of datasets on which the proposed SADD outperforms other discretization methods, and $v$ is the number of datasets on which the performance gain is statistically significant with the significance level of $p = 0.05$.}
 \resizebox{1\columnwidth}{!}{
	\begin{tabular}{@{}ccccccccc@{}}
		\toprule
		 & MDLP   & CAIM   & CACC   & ChiMerge & EW     & EF & PKID &FFD     \\ \midrule
		SADD      & 31(19) & 31(13) & 30(15) & 31(25)   & 27(18) & 30(21) &27(18)&26(18)\\ \bottomrule
	\end{tabular}
 }
	\label{sum}
\end{table}

\subsection{Analysis and Discussion of Proposed SADD against MDLP}
\begin{table}[htp]
	\centering
	\caption{\scriptsize{The performance gain (PG) of the proposed SADD over MDLP~\citep{fayyad1993multi} on 31 datasets, and the number of features discretized into various number of intervals (Num) by the two methods. Obviously many features are discretized into one interval by MDLP~\citep{fayyad1993multi}, which results in a huge information loss. In contrast, the proposed SADD discretizes features into more intervals and retains more discriminant information.}}
	\resizebox{.85\columnwidth}{!}{
		\begin{tabular}{|c|c|c|c|c|c|c|c|c|c|c|c|c|c|}
			\hline
			\multicolumn{1}{|l|}{} & \multirow{2}{*}{PG(\%)} & \multicolumn{6}{c|}{Proposed SADD} & \multicolumn{6}{c|}{MDLP}   \\ \cline{1-1} \cline{3-14} 
			\diagbox[width=90pt,height=30pt]{\scriptsize{Dataset}}{\scriptsize{Num}}               &                                      & 1    & 2   & 3   & 4   & 5   & \textgreater{}5  & 1  & 2  & 3  & 4  & 5  & \textgreater{}5 \\ \hline
   Iris	& 3.33                                 & -    & -   & 1   & 2   & 1   & -   & -   & 2  & 2  & -  & -  & -  \\ \hline
Parkinson	& 4.55                                 & 1    & 9   & 6   & 4   & 3   & -   & 2   & 18 & 3  & -  & -  & -  \\ \hline
Seeds	& 3.81                                 & -    & -   & 1   & 2   & 4   & -   & -   & 3  & -  & 3  & 1  & -  \\ \hline
Glass	& 2.25                                 & 1    & -   & 1   & 1   & 1   & 6   & 3   & 3  & 2  & 2  & -  & -  \\ \hline
Heart	& 0.00                                 & 1    & 5   & -   & -   & -   & -   & 3   & 3  & -  & -  & -  & -  \\ \hline
Ecoli	& 3.52                                 & 2    & 1   & 1   & 1   & 1   & 2   & 2   & 3  & 3  & -  & -  & -  \\ \hline
Bupa	& 12.49                                & 1    & 3   & 2   & -   & -   & -   & 5   & 1  & -  & -  & -  & -  \\ \hline
Ionophere	& 1.10                                 & -    & -   & 2   & 3   & 7   & 20  & 1   & 1  & 12 & 2  & 12 & 4  \\ \hline
Movement	& 15.67                                & -    & 1   & 5   & 10  & 7   & 67  & 21  & 25 & 29 & 15 & -  & -  \\ \hline
ILPD	& 2.23                                 & 2    & 4   & 2   & 2   & -   & -   & 5   & 5  & -  & -  & -  & -  \\ \hline
Breast	& 0.29                                 & -    & 1   & 2   & 3   & 1   & 2   & -   & 1  & 5  & 2  & -  & 1  \\ \hline
Pima	& 3.13                                 & -    & 3   & 1   & 4   & -   & -   & 2   & 4  & 1  & 1  & -  & -  \\ \hline
Vowel	& 16.67                                & -    & 1   & -   & -   & -   & 9   & 1   & -  & 4  & 3  & -  & 2  \\ \hline
Biodegradation	& 1.04                                 & 4    & 10  & 14  & 7   & 2   & 4   & 8   & 15 & 15 & 2  & 1  & -  \\ \hline
Mice Protein	& 4.08                                 & 2    & 7   & -   & 5   & 13  & 55  & 4   & 25 & 16 & 15 & 9  & 13 \\ \hline
Yeast	& 2.63                                 & 2    & 3   & -   & 3   & 2   & -   & 4   & 2  & 3  & 1  & -  & -  \\ \hline
Mfeat-fac	&6.25 								&115 &84  &53  &33  &26  &39  &191  &91  &37  &16  &5  &2\\ \hline
Cardio	& 1.51                                 & 1    & 2   & 2   & 2   & 1   & 15  & 2   & 2  & 5  & 4  & 2  & 8  \\ \hline
Madelon	& 2.73                                 & 484  & 7   & 5   & 2   & 2   & -   & 487 & 9  & 1  & 3  & -  & -  \\ \hline
Spambase	& 0.54                                 & 2    & 17  & 17  & 11  & 3   & 7   & 2   & 29 & 15 & 5  & 4  & 2  \\ \hline
Wave	& 0.52                                 & 2    & -   & 1   & -   & 1   & 17  & 2   & -  & 2  & 3  & 2  & 12 \\ \hline
Wall-Following	& 1.72                                 & -    & -   & -   & -   & -   & 24  & -   & -  & -  & -  & -  & 24 \\ \hline
Page-block	& 0.31                                 & -    & -   & -   & -   & 1   & 9   & -   & -  & -  & -  & 2  & 8  \\ \hline
Opdigit	& 0.27                                 & 7    & 7   & 3   & 3   & 8   & 36  & 7   & 9  & 3  & 11 & 18 & 16 \\ \hline
Satellite	& 0.31                                 & -    & -   & -   & -   & -   & 36  & -   & -  & -  & -  & -  & 36 \\ \hline
Wine	& 1.23                                 & 1    & -   & 2   & 1   & 3   & 4   & 1   & 1  & 3  & 4  & 2  & -  \\ \hline
Musk    & 1.23                                 & 1 & 1   & 3   & 5   & -   & 156   & 3   & 3  & 8  & 7  & 10  & 135  \\ \hline
Anuran	& 0.69                                 & -    & 1   & -   & -   & -   & 21  & -   & 1  & -  & -  & -  & 21 \\ \hline
Pendigit	& 0.33                                 & -    & -   & -   & -   & -   & 16  & -   & -  & -  & -  & -  & 16 \\ \hline
Magic	& 0.46                                 & -    & -   & -   & 1   & 2   & 7   & -   & -  & -  & 1  & 2  & 7  \\ \hline
IndoorLoc	&1.65 								&0 &2 &5 &13 &32 &164 &1 &8 &59 &60 &58 &30 \\ \hline
		\end{tabular}
	}
	\label{interval}
\end{table}
The proposed SADD is based on the MDLP~\citep{fayyad1993multi} method but performs significantly better. In Section~\ref{sec:SADD}, we show that in theory the proposed SADD could effectively prevent the information loss and indeed the proposed SADD does significantly outperform MDLP~\citep{fayyad1993multi} on most datasets as shown in the previous subsection. To analyze the underlying reasons why the proposed SADD performs better than MDLP~\citep{fayyad1993multi}, Table~\ref{interval} summarizes the number of features discretized into the various number of intervals by MDLP~\citep{fayyad1993multi} and the proposed SADD, respectively, and the performance gain of SADD on each dataset compared with MDLP~\citep{fayyad1993multi}. For example, for six features of the ``Bupa'' dataset, five features are discretized into one interval and one is discretized into two intervals by MDLP~\citep{fayyad1993multi}, which leads to a huge information loss. When an attribute is discretized into one interval, the attribute values for all classes are the same. As a result, this attribute can not be used to differentiate different classes and hence the discriminant information residing in the attribute is totally lost. By utilizing the proposed discretization, most features are discretized into more intervals, and hence the discriminant information is better preserved. 

As shown in Table~\ref{interval}, the proposed SADD performs best on almost all datasets. On datasets with little discriminant information loss during discretization such as ``Anuran", ``Magic", ``Page-Block", ``Pendigit" and ``Satellite", the performance gains on these datasets are relatively small. On datasets with significant information loss such as ``Bupa", ``Mice Protein", ``Movement", ``Parkinson" and ``Vowel", the performance gains are high, e.g., the performance gains on ``Bupa", ``Movement" and ``Vowel" are more than 10\%. Table~\ref{interval} clearly demonstrates that the proposed SADD could well address the problem of discriminant information loss in MDLP. It generates a proper number of intervals to preserve the discrimination power of classification algorithms, and at the same time retain the generalization capability.

\rev{To analyze the performance gains of the proposed SADD over MDLP~\citep{fayyad1993multi} on different datasets, we group the datasets according to the instance size and the feature size, respectively. 1) In terms of instance size, the proposed SADD greatly enhances the discrimination power of NB classifier on both relatively small datasets (\# of Inst. $\leq$ 1000) and relatively large datasets (\# of Inst. $>$ 1000), with an average performance gain of 5.31\% and 1.53\%, respectively. The performance gains on relatively small datasets are more significant because MDLP is more likely to stop the top-down split in the early stages for small datasets. As shown in Fig.~\ref{curve2}, the large threshold for a small $N$ may cause the early stop of MDLP, and hence lead to a significant performance drop of MDLP, while the proposed SADD well tackles this problem by utilizing a significantly smaller threshold. 2) In terms of feature size, the proposed SADD greatly enhances the discrimination power of NB classifier on both datasets with relatively few features (\# of Feat. $\leq$ 50) and datasets with relatively many features (\# of Feat. $>$ 50), with an average performance gain of 2.79\% and 4.05\%, respectively. The performance gains on datasets with more features are more significant because naive Bayes could aggregate the discriminant information gains of more features, resulting in better performance.}

\subsection{Comparisons to State-of-the-art Discretization Methods for Semi-supervised Learning}
\rev{To evaluate the proposed SADD in a more challenging scenario for semi-supervised learning, we follow the experimental setting in \citep{lai2022adaptive,lai2022semi}, where 40\% of training samples are randomly selected as labeled data and the rest are treated as unlabeled data. The proposed SADD utilizes pseudo-labeling techniques to label the unlabeled training data for discretization. 
The comparisons to four supervised discretization methods under this setting are summarized in Table~\ref{ablation-semi}. The results of unsupervised discretization methods such as Equal-W~\citep{kurgan2004caim}, Equal-F~\citep{kurgan2004caim}, PKID~\citep{yang2001proportional} and FFD~\citep{yang2009discretization} may refer back to Table~\ref{res}. 
As shown in Table~\ref{ablation-semi}, the proposed SADD achieves an improvement of 3.55\% on average compared with MDLP~\citep{fayyad1993multi}. Compared with CAIM~\citep{kurgan2004caim}, CACC~\citep{tsai2008discretization} and ChiMerge~\citep{kerber1992chimerge}, the proposed SADD obtains the improvements of 1.53\%, 2.24\% and 4.90\%, respectively.}

\begin{table}[!hbt]

\centering
\caption{\scriptsize{\rev{Comparisons between the proposed SADD and other supervised discretization methods where 40\% of training samples are labeled data and the rest are unlabeled data. The proposed SADD obtains an average performance gain of 3.55\% compared with MDLP ~\citep{fayyad1993multi}.}}}
\resizebox{0.85\columnwidth}{!}{
\begin{tabular}{cccccc}
\toprule
Dataset        & SADD                 & MDLP       & CAIM                & CACC                & ChiMerge            \\ \midrule
Iris                          & \textBF{96.00±4.66}  & 95.33±5.49 & 92.00±6.89          & 92.67±5.84  & 76.67±11.86 \\
Parkinson                     & \textBF{83.07±7.18}  & 79.88±5.66 & 81.41±9.12          & 79.96±7.18  & 78.60±6.51  \\
Seeds                         & \textBF{89.05±5.04}  & 88.57±6.02 & 87.62±6.02          & 87.62±7.17  & 81.90±7.03  \\
Glass                         & \textBF{69.62±12.95} & 59.92±8.72 & 68.71±11.80         & 66.85±10.21 & 64.37±10.43 \\
Heart                         & \textBF{84.81±8.63}  & 84.44±9.04 & 84.44±8.69          & 82.96±9.43  & 82.96±9.27  \\
Ecoli                         & \textBF{86.13±6.34}  & 83.13±6.30 & 82.80±4.37          & 83.68±6.18  & 81.88±6.54  \\
Bupa                          & \textBF{64.03±3.99}  & 55.63±7.19 & 62.56±5.86          & 63.99±6.69  & 62.01±8.79  \\
Ionophere                     & \textBF{90.35±3.92}  & 90.09±6.38 & 89.49±5.75          & 88.62±4.95  & 78.06±5.69  \\
Movement                      & \textBF{72.04±7.72}  & 38.74±7.92 & 66.90±12.33         & 70.86±8.55  & 70.50±6.20  \\
ILPD                          & \textBF{68.44±4.22}  & 65.34±5.12 & 67.57±5.80          & 66.89±5.73  & 64.31±3.00  \\
Breast                        & \textBF{97.57±2.03}  & 97.28±1.43 & 97.14±1.79          & 97.14±1.79  & 94.28±2.94  \\
Pima                          & \textBF{77.08±3.95}  & 74.73±4.20 & 74.86±5.16          & 74.87±2.89  & 71.34±5.82  \\
Vowel                         & \textBF{65.05±4.37}  & 47.17±8.36 & 63.94±6.95          & 52.73±6.96  & 60.61±4.39  \\
Biodegradation                & \textBF{81.23±3.43}  & 77.82±3.85 & 80.38±3.16          & 81.23±2.98  & 77.82±3.58  \\
Mice Protein                  & \textBF{94.62±3.75}  & 94.53±2.62 & 93.23±3.02          & 91.66±3.18  & 92.12±3.56  \\
Yeast                         & \textBF{59.85±4.28}  & 58.03±4.29 & 57.96±3.19          & 55.81±3.64  & 57.02±4.97  \\
Mfeat-fac                       & \textBF{93.95±2.28}  & 92.85±1.90 & 93.85±2.25          & 93.60±2.54  & 93.15±2.17  \\
Cardio                        & 80.10±1.76           & 78.03±2.35 & \textBF{80.20±2.66} & 79.78±1.62  & 77.38±1.70  \\
Madelon                       & \textBF{63.31±4.03}  & 60.42±3.89 & 58.15±3.88          & 50.00±0.00  & 59.31±4.09  \\
Spambase                      & \textBF{90.24±1.58}  & 90.09±1.58 & 89.70±1.58          & 90.02±1.50  & 89.26±1.10  \\
Wave                          & \textBF{80.24±1.23}  & 79.76±1.28 & 79.92±1.08          & 74.80±0.97  & 78.72±1.24  \\
Wall-Following                & \textBF{90.16±1.24}  & 88.36±1.36 & 86.82±2.13          & 88.64±1.80  & 71.28±1.71  \\
Page-Block                    & \textBF{93.92±1.34}  & 93.64±1.31 & 93.09±1.15          & 93.44±1.57  & 91.54±1.05  \\
Opdigit                       & \textBF{92.46±0.65}  & 91.57±0.78 & 92.40±0.70          & 92.28±0.67  & 92.01±0.78  \\
Satellite                     & \textBF{82.44±1.41}  & 81.79±1.43 & 81.97±1.43          & 82.07±1.28  & 79.60±1.54  \\
Wine                          & 49.38±1.95           & 48.75±1.02 & \textBF{50.01±1.86} & 49.47±2.03  & 48.52±1.52  \\
Musk                          & \textBF{91.94±0.77}  & 90.27±1.00 & 86.22±2.00          & 89.80±0.78  & 78.18±1.94  \\
Anuran                        & \textBF{90.40±1.28}  & 89.58±1.43 & 89.28±1.46          & 89.21±1.57  & 81.88±1.40  \\
Pendigit                      & \textBF{88.27±0.75}  & 87.35±0.69 & 87.67±0.55          & 88.12±0.69  & 87.02±0.82  \\
Magic                         & \textBF{76.95±0.47}  & 76.77±0.77 & 75.74±0.48          & 76.04±0.58  & 73.36±0.92  \\
IndoorLoc                     & \textBF{62.91±1.20}  & 55.68±0.89 & 62.24±0.93          & 61.50±1.12  & 58.28±1.06  \\ \midrule
\textBF{AVG}                  & 80.83                & 77.28      & 79.30               & 78.59       & 75.93                        \\ \bottomrule
\end{tabular}
}
\label{ablation-semi}
\end{table}

\rev{Table~\ref{sum2} summarizes the results for statistical significance tests. Among 31 datasets, the proposed SADD outperforms MDLP~\citep{fayyad1993multi}, CAIM~\citep{kurgan2004caim}, CACC~\citep{tsai2008discretization} and ChiMerge~\citep{kerber1992chimerge} on 31, 29, 30 and 31 datasets respectively, among which 15, 9, 11 and 25 are statistically significant.}
\begin{table}[!ht]
	\centering
	\caption{\rev{Summary of statistical significance tests of the proposed SADD over supervised discretization methods when 40\% of training samples are labeled data while the rest are unlabeled.}}
 \resizebox{0.6\columnwidth}{!}{
	\begin{tabular}{@{}ccccc@{}}
		\toprule
		   & MDLP   & CAIM   & CACC   & ChiMerge      \\ \midrule
		SADD      & 31(15) & 29(9) & 30(11) & 31(25)   \\ \bottomrule
	\end{tabular}
 }
	\label{sum2}
\end{table}

\subsection{Comparisons to State-of-the-art Naive Bayes Classifiers}
\rev{The proposed SADD can be integrated with not only regularized naïve Bayes, but also other naïve Bayes classifiers. To demonstrate the performance gain brought by the proposed SADD, we integrate it with CAWNB~\citep{jiang2019class}, WANBIA~\citep{zaidi2013alleviating}, AIWNB$^E$~\citep{zhang2021attribute} and AIWNB$^L$~\citep{zhang2021attribute} and RNB~\citep{RNB2020shihe} resulting in CAWNB+, WANBIA+, AIWNB$^E$+, AIWNB$^L$+ and RNB+ respectively. Note that the MDLP discretization scheme was previously utilized in these NB classifiers. The comparison results are summarized in Table~\ref{res2}. As shown in Table~\ref{res2}, the proposed SADD has greatly enhanced the performance of these state-of-the-art NB classifiers, and the performance gains on CAWNB, WANBIA, AIWNB$^E$, AIWNB$^L$ and RNB are 1.99\%, 1.82\%, 2.71\%, 2.16\% and 2.16\%, respectively. These results demonstrate that the proposed SADD discretization scheme can be seamlessly integrated with various naïve Bayes classifiers and significantly improve their performance.}

\begin{table}[!hp]
    	\caption{\footnotesize{\rev{Summary of performance gain brought by the proposed SADD on state-of-the-art naive Bayes classifiers, where CAWNB+, WANBIA+, AIWNB$^E$+, AIWNB$^L$+ and RNB+ utilize the proposed SADD discretization scheme and others utilize MDLP~\citep{fayyad1993multi}.}}}

	\resizebox{.99\columnwidth}{!}{
  \begin{tabular}{ccccccccccc}
  \toprule
Dataset        & CAWNB      & CAWNB+              & WANBIA              & WANBIA+     & AIWNB$^E$   & AIWNB$^E$+           & AIWNB$^L$            & AIWNB$^L$+           & RNB                 & RNB+                 \\ \midrule
Iris           & 93.33±5.96 & 96.00±3.44          & 93.33±5.96          & 96.67±3.51  & 92.67±6.29 & 96.00±4.66          & 92.67±6.29          & 96.67±3.51          & 93.33±5.96          & \textBF{96.67±3.51}  \\
Parkinson      & 84.68±5.88 & 89.79±6.48          & 85.76±6.59          & 90.34±6.06  & 79.46±5.24 & 84.52±6.68          & 81.55±6.19          & 86.63±5.54          & 85.23±6.04          & \textBF{90.34±6.06}  \\
Seeds          & 90.00±6.88 & 91.90±4.52          & 90.00±5.81          & 92.38±4.60  & 87.62±4.36 & 90.48±6.35          & 87.62±4.36          & 91.90±5.04          & 89.52±7.00          & \textBF{92.38±4.02}  \\
Glass          & 73.85±3.51 & 73.72±7.72          & 71.13±8.30          & 72.42±5.51  & 74.28±6.86 & 74.29±5.92          & \textBF{75.28±7.75} & 74.81±4.74          & 71.06±4.22          & 73.77±7.55           \\
Heart          & 77.41±9.86 & 83.70±9.43          & 84.07±10.08         & 83.33±10.80 & 83.70±8.64 & 83.33±9.76          & 83.70±9.10          & 83.33±9.11          & \textBF{84.07±9.81} & 83.33±10.66          \\
Ecoli          & 83.38±3.39 & 85.41±3.81          & 82.51±3.79          & 84.89±6.12  & 82.23±4.92 & 84.84±4.43          & 82.23±4.92          & 84.55±5.93          & 83.39±3.34          & \textBF{85.41±3.81}  \\
Bupa           & 53.27±9.52 & 62.51±11.86         & 53.27±9.52          & 62.51±11.86 & 42.02±0.84 & 62.24±9.24          & 42.02±0.84          & 61.67±9.71          & 53.27±9.52          & \textBF{62.51±11.86} \\
Ionophere      & 89.23±5.10 & 90.63±4.56          & \textBF{92.94±6.17} & 91.76±5.38  & 89.52±4.96 & 90.65±5.08          & 90.37±4.83          & 92.08±5.25          & 91.81±6.03          & 90.35±5.95           \\
Movement       & 68.42±5.27 & 74.87±4.31          & 67.16±3.44          & 74.94±8.04  & 64.63±4.83 & 75.39±7.60          & 67.12±4.33          & 75.28±6.23          & 65.63±6.88          & \textBF{76.83±5.32}  \\
ILPD           & 67.57±4.52 & 69.97±4.66          & 67.57±4.52          & \textBF{69.98±4.72}  & 66.54±4.43 & 67.74±4.29          & 66.89±3.90          & 67.92±3.40          & 67.92±4.35          & 69.63±4.48  \\
Breast         & 95.85±1.49 & 96.28±2.04          & 96.28±1.59          & 96.42±2.64  & 97.28±1.20 & \textBF{97.57±1.66} & 96.99±1.50          & 97.42±1.89          & 96.42±1.60          & 96.42±1.55           \\
Pima           & 75.12±5.58 & 76.17±4.49          & 74.21±4.76          & 76.17±4.08  & 73.69±4.34 & 74.60±5.53          & 73.82±4.64          & 75.12±5.32          & 74.86±5.41          & \textBF{76.17±4.31}  \\
Vowel          & 61.11±4.99 & 76.26±5.54          & 61.62±4.78          & 76.77±5.59  & 59.70±4.18 & 76.97±4.06          & 66.97±5.11          & 72.63±4.80          & 60.91±4.47          & \textBF{76.97±4.97}  \\
Biodegradation & 84.64±2.32 & 85.21±3.01          & 85.12±2.44          & 85.69±3.18  & 81.13±2.75 & 82.84±2.05          & 81.61±2.38          & 83.69±1.76          & 85.21±2.33          & \textBF{85.78±3.26}  \\
Mice Protein   & 98.89±1.29 & 99.82±0.39          & 99.63±0.62          & 99.72±0.44  & 97.32±1.32 & 99.45±0.77          & 98.33±1.15          & 99.54±0.65          & 99.63±0.62          & \textBF{99.91±0.29}  \\
Yeast          & 57.56±3.94 & 59.37±3.44          & 56.75±4.07          & 59.38±4.64  & 57.15±3.46 & 59.17±4.41          & 57.15±3.80          & 59.32±4.36          & 57.29±3.99          & \textBF{59.38±3.97}  \\
Mfeat-fac      & 93.85±2.14 & 94.80±1.90          & 95.55±1.40          & 95.90±1.07  & 94.20±1.77 & 95.10±1.54          & 95.15±1.49          & \textBF{95.95±1.40} & 95.40±1.76          & 95.65±0.91           \\
Cardio         & 88.66±1.49 & 88.94±1.88          & 88.66±1.56          & 88.42±1.89  & 80.53±1.42 & 81.33±2.16          & 82.64±1.32          & 82.79±1.62          & 88.70±1.87          & \textBF{89.13±1.50}  \\
Madelon        & 63.81±3.47 & 64.96±3.63          & 63.35±3.68          & 64.88±3.92  & 61.92±3.34 & 64.65±3.99          & 61.92±3.34          & 64.62±3.96          & 62.96±3.52          & \textBF{65.15±3.11}  \\
Spambase       & 94.11±0.74 & \textBF{94.26±1.08} & 93.78±0.92          & 93.89±1.28  & 89.94±1.25 & 90.35±1.78          & 90.16±1.27          & 90.53±1.85          & 93.94±1.19          & 93.96±1.06           \\
Wave           & 84.36±1.22 & 84.86±1.44          & 83.88±1.52          & 84.16±1.54  & 80.08±1.05 & 80.48±1.09          & 80.74±1.28          & 81.22±0.94          & 84.30±1.56          & \textBF{85.08±1.19}  \\
Wall-Following & 96.52±1.57 & 96.56±1.75          & 97.42±0.57          & 97.49±0.63  & 90.95±1.31 & 91.90±0.83          & 93.49±0.84          & 93.53±0.59          & \textBF{97.53±0.65}          & 97.43±0.77  \\
Page-Block     & 96.38±0.65 & \textBF{96.60±0.69} & 96.04±0.83          & 96.31±0.74  & 93.02±1.30 & 93.11±1.32          & 93.79±1.07          & 94.37±1.30          & 96.33±0.87          & 96.55±0.78           \\
Opdigit        & 94.73±0.94 & 94.95±1.18          & 93.45±1.04          & 93.83±1.03  & 92.30±0.43 & 92.60±0.64          & 93.11±0.37          & 93.31±0.70          & 95.23±0.81          & \textBF{95.77±0.72}  \\
Satellite      & 84.40±1.01 & 84.48±1.40          & 84.52±0.76          & 85.00±0.93  & 81.69±1.26 & 81.88±1.62          & 85.33±1.00          & 85.72±1.33          & 85.81±0.92          & \textBF{86.37±0.91}  \\
Wine           & 51.70±1.21 & 52.20±1.36          & 53.19±1.37          & 53.63±1.94  & 48.99±1.38 & 50.33±1.03          & 50.64±1.35          & 51.50±1.26          & 53.36±1.66          & \textBF{53.70±1.10}  \\
Musk & 97.24±0.73 & \textBF{97.33±0.44} & 96.23±0.87 & 96.15±0.40 & 92.91±0.86 & 92.89±0.81 & 93.79±0.93 & 93.66±0.64 & 95.89±0.68 & 97.04±0.71 \\
Anuran         & 95.41±0.68 & 95.55±0.84          & 94.66±0.57          & 95.11±0.60  & 88.87±1.07 & 89.46±1.18          & 92.93±1.08          & 93.58±1.08          & 95.44±0.68          & \textBF{95.97±0.75}  \\
Pendigit       & 93.06±0.42 & 93.76±0.62          & 89.71±0.76          & 90.04±0.76  & 88.72±1.11 & 89.24±1.22          & 93.41±0.68          & 93.72±0.71          & 93.15±0.35          & \textBF{93.81±0.71}  \\
Magic          & 83.43±0.75 & \textBF{83.61±0.65}          & 82.40±0.63          & 82.47±0.77  & 79.32±0.49 & 79.81±0.54          & 80.18±0.62          & 80.71±0.38          & 83.40±0.72          & 83.41±0.83  \\
IndoorLoc      & \textBF{87.09±1.53} & 86.30±3.92          & 86.30±0.66          & 86.52±0.76  & 65.27±1.26 & 68.56±1.13          & 68.08±0.96           & 68.66±1.23           & 83.59±2.99          & 86.64±0.96  \\ \midrule
\textBF{AVG}           & 82.55 & 84.54 & 82.60 & 84.42 & 79.28 & 81.99 & 80.63 & 82.79 & 82.73 & 84.89               \\ \bottomrule
\end{tabular}
	}
	\label{res2}
\end{table}

\rev{Table~\ref{sum1} summarizes the statistical significance tests of the proposed SADD over MDLP on various NB classifiers. By utilizing the proposed SADD discretization scheme, CAWNB+, WANBIA+, AIWNB$^E$+, AIWNB$^L$+ and RNB+ achieve the higher classification performance on most of the datasets than their counterparts, CAWNB, WANBIA, AIWNB$^E$, AIWNB$^L$ and RNB, respectively, among which the results on 9, 6, 14, 9, and 14 datasets are statistically significant.}

\begin{table}[!ht]
	\caption{{Summary of statistical significance tests of the proposed SADD over MDLP~\citep{fayyad1993multi} on various NB classifiers. For each entry $u(v)$, $u$ is the number of datasets on which CAWNB+, WANBIA+, AIWNB$^E$+, AIWNB$^L$+ and RNB+ outperform their counterparts, and $v$ is the number of datasets on which the performance gain is statistically significant with the significance level of $p = 0.05$.}}
	\centering
 \resizebox{.99\columnwidth}{!}{
	\begin{tabular}
 {|c|c|c|c|c|}
		\toprule
		      CAWNB+ vs. & WANBIA+ vs.& AIWNB$^E$+ vs. & AIWNB$^L$+ vs. & RNB+ vs. \\     
        	 CAWNB  &  WANBIA &  AIWNB$^E$ & AIWNB$^L$ &  RNB \\ \midrule 
		      28(9) & 27(6) & 29(14)  & 19(9)& 28(14)  \\ \bottomrule
	\end{tabular}
 }
	\label{sum1}
\end{table}

\section{Conclusion}
\label{sec:conclu}
In this paper, we aim to design a discretization and classification framework to balance the generalization capability and discrimination power, during both data discretization and classification. We find that a popular discretization scheme, MDLP, often results in an early stop during the top-down discretization, which leads to a huge information loss. To address this problem, we propose a semi-supervised adaptive discriminative discretization (SADD) method, which utilizes the pseudo-labeling technique to make full use of the discriminant information residing in both labeled and unlabeled data. Furthermore, an adaptive discriminative discretization scheme is designed to resolve the problem of huge information loss in MDLP. In such a way, the proposed SADD retains the discriminant information for the classifier while preserving its generalization ability. Besides, the proposed RNB+ well balances the generalization ability and discrimination power during both data discretization and feature weighting. Experimental results on 31 machine-learning datasets demonstrate that the proposed SADD significantly outperforms all compared discretization methods and the proposed RNB+ significantly outperforms other state-of-the-art NB classifiers.

\section*{Declaration of Competing Interest}
The authors declare that they have no known competing financial interests or personal relationships that could have appeared to influence the work reported in this paper.
\section*{Acknowledgement}
This work was supported in part by the National Natural Science Foundation of China under Grant 72071116, and in part by the Ningbo Municipal Bureau Science and Technology under Grants 2019B10026 and 2017D10034.
\small
\bibliography{Template}
\end{document}